\documentclass[10pt,twocolumn,letterpaper]{article}

\usepackage[]{cvpr}      

\usepackage{graphicx}
\usepackage{amsmath}
\usepackage{amssymb}
\usepackage{booktabs}
\usepackage{multirow}
\usepackage{color}
\usepackage{colortbl}
\definecolor{light}{rgb}{0.9,0.9,0.9}
\definecolor{com}{rgb}{.208, .688, .556}
\definecolor{incom}{rgb}{.864, .368, .0}
\usepackage[accsupp]{axessibility}
\usepackage{pythonhighlight}
\definecolor{cvprblue}{rgb}{0.21,0.49,0.74}
\usepackage[pagebackref,breaklinks,colorlinks,citecolor=cvprblue]{hyperref}

\def\paperID{4046} 
\def\confName{CVPR}
\def\confYear{2024}

\title{Commonsense Prototype for Outdoor Unsupervised 3D Object Detection}

\author{Hai Wu$^1$ \quad Shijia Zhao$^{1}$ \quad Xun Huang$^{1}$ \quad Chenglu Wen$^1$\thanks{Corresponding author} \quad Xin Li$^2$ \quad Cheng Wang$^1$ \\
$^1$Fujian Key Laboratory of Sensing and Computing for Smart Cities, Xiamen University\\ \quad $^2$ Section of Visual Computing and Interactive Media, Texas A\&M University\\
}

\begin{document}

\maketitle
\begin{abstract}
The prevalent approaches of unsupervised 3D object detection follow cluster-based pseudo-label generation and iterative self-training processes. However, the challenge arises due to the sparsity of LiDAR scans, which leads to pseudo-labels with erroneous size and position, resulting in subpar detection performance. To tackle this problem, this paper introduces a \textbf{C}ommonsense \textbf{P}rototype-based \textbf{D}etector, termed \textbf{CPD}, for unsupervised 3D object detection.  CPD first constructs Commonsense Prototype (CProto) characterized by high-quality bounding box and dense points, based on commonsense intuition. Subsequently, CPD refines the low-quality pseudo-labels by leveraging the size prior from CProto. Furthermore, CPD enhances the detection accuracy of sparsely scanned objects by the geometric knowledge from CProto. CPD outperforms state-of-the-art unsupervised 3D detectors on Waymo Open Dataset (WOD), PandaSet, and KITTI datasets by a large margin. Besides, by training CPD on WOD and testing on KITTI, CPD attains \textbf{90.85\%} and \textbf{81.01\%} 3D Average Precision on easy and moderate car classes, respectively. These achievements position CPD in \textbf{close proximity to fully supervised detectors}, highlighting the significance of our method. The code will be available at \url{https://github.com/hailanyi/CPD}.
\end{abstract}
    
\section{Introduction}
\label{sec:intro}
Autonomous driving requires reliable detection of 3D objects (e.g. vehicle and cyclist) in urban scenes for safe path planning and navigation. Thanks to the power of neural networks, numerous studies have developed high-performance 3D detectors through fully supervised approaches\cite{Voxel-RCNN, SFD, BEVFusion, CasA, TED, VirConv}. However, these models heavily depend on human annotations from diverse scenes to guarantee their effectiveness across various scenarios.
This data labeling process is typically laborious and time-consuming, limiting the wide deployment of detectors in practice~\cite{MODEST}. 

\begin{figure}[t]
  \centering
   \includegraphics[width=1\linewidth]{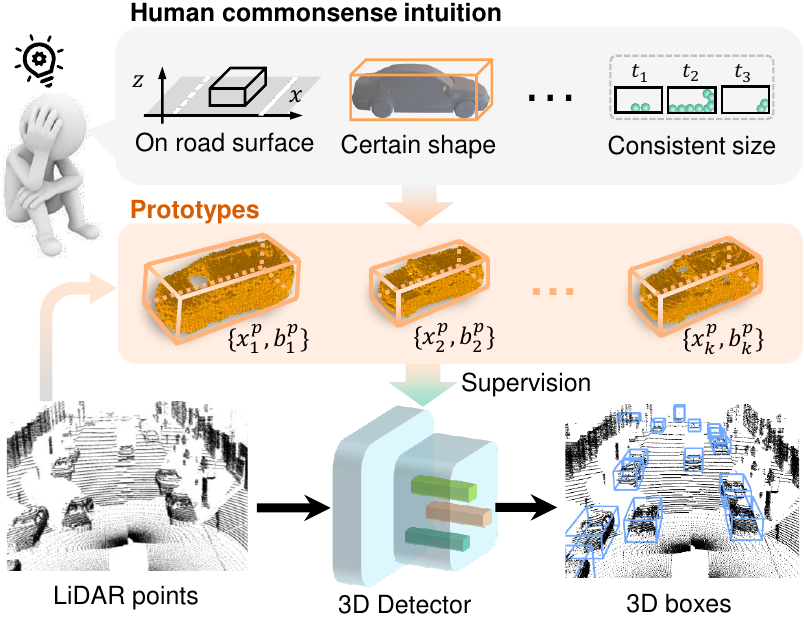}
   \caption{Illustration of commonsense prototypes for unsupervised 3D object detection in autonomous driving scenes.}
   \label{fig:idea}
\end{figure}

Several studies have explored approaches to reduce labeling requirements by weakly supervised learning~\cite{NoiseDet,3DIoUMatch, SESS}, decreasing the label cost by over 80\%. Notably, the objects within a 3D scene exhibit distinguishable attributes and can be easily identified through certain commonsense reasoning (see Fig.~\ref{fig:idea}). 
For example, the objects are usually located on the ground surface with a certain shape; the object sizes are fixed across frames. This insight has prompted us to develop an \textit{unsupervised 3D detector} that operates without using human annotations.

\begin{figure}[t]
  \centering
   \includegraphics[width=1\linewidth]{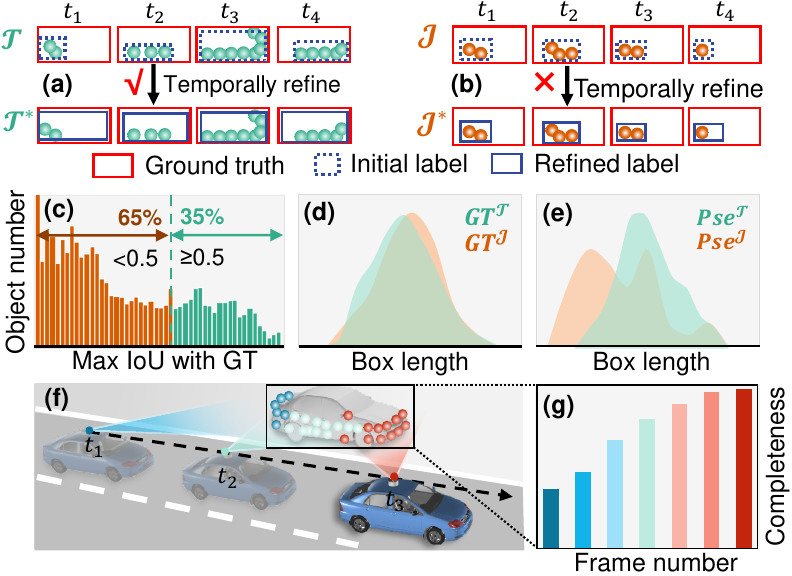}
   \caption{Illustration and statistics of complete and incomplete objects on WOD~\cite{Waymo} validation set (large enough to demonstrate the general problem). (a) Pseudo-labels of \textit{\textcolor{com}{complete object $\mathcal{T}$}} are refined by temporal consistency. (b) Pseudo-labels of \textit{\textcolor{incom}{incomplete object $\mathcal{J}$}} fail to be refined by temporal consistency. (c) 65\% objects lack full scan coverage and generate inaccurate pseudo-labels ( Max IoU (Intersection over Union) $<$ 0.5 with GT (Ground Truth)). (d) The vehicle GT of \textit{\textcolor{com}{complete object $GT^\mathcal{T}$}} and \textit{\textcolor{incom}{incomplete object $GT^\mathcal{J}$}} have similar size distributions. (e) The pseudo-label of \textit{\textcolor{com}{complete object $Pse^\mathcal{T}$}} and \textit{\textcolor{incom}{incomplete object $Pse^\mathcal{J}$}} have different size distributions. (f)(g) The nearby stationary objects are with high completeness in consecutive frames.}
   \label{fig:motivation}
\end{figure}

In recent years, traditional methods leveraged ground removal~\cite{GroundRemoval} and clustering technique~\cite{2013shape} for unsupervised 3D object detection. However, these methods often struggle to achieve satisfactory performance due to the sparsity and occlusion of objects in 3D scenes. Advanced methods create initial pseudo-labels from point cloud sequences by clustering and bootstrap a good detector by iteratively training a deep network~\cite{OYSTER}. Nevertheless, the sparse and view-limited nature of LiDAR scanning leads to pseudo-labels with inaccurate sizes and positions, misleading the network convergence and resulting in suboptimal detection performance. A subset of objects, denoted as \textit{\textcolor{com}{complete objects $\mathcal{T}$}}, benefit from having at least one complete scan across the entire point cloud sequence, allowing their pseudo-labels to be refined through temporal consistency~\cite{OYSTER} (see  Fig.~\ref{fig:motivation} (a)). However, the majority of objects (e.g. 65\% on WOD~\cite{Waymo}, as shown in Fig.~\ref{fig:motivation} (c)), termed \textit{\textcolor{incom}{incomplete objects $\mathcal{J}$}}, lack full scan coverage (see Fig.~\ref{fig:motivation} (b)), and cannot be recovered by temporal consistency. 

To tackle this issue, this paper proposes a Commonsense Prototype-based Detector, termed \textbf{CPD}, for \textit{unsupervised 3D object detection}. CPD is built upon two key insights: (1) The ground truth of intra-class objects keeps a similar size (length, width, and height) distribution between incomplete objects and complete objects (see  Fig.~\ref{fig:motivation} (d)). (2) The nearby stationary objects are very complete in consecutive frames and can be recognized accurately by commonsense intuition (see  Fig.~\ref{fig:motivation} (f)(g)). 
Our idea is to construct a Commonsense Prototype (\textbf{CProto}) set representing accurate geometry and size from complete objects to refine the pseudo-labels of incomplete objects and improve the detection accuracy. 
To this end, we first design an unsupervised Multi-Frame Clustering (\textbf{MFC}) method that yields high-recall initial pseudo-labels. Subsequently, we introduce an unsupervised Completeness and Size Similarity (\textbf{CSS}) score that selects high-quality labels to construct the CProto set. Furthermore, we design a CProto-constrained Box Regularization (\textbf{CBR}) method to refine the pseudo-labels by incorporating the size prior from CProto. In addition, we develop CProto-constrained Self-Training (\textbf{CST}) that improves the detection accuracy of sparsely scanned objects by the geometry knowledge from CProto. 

The effectiveness of our design is verified by experiments on widely used WOD~\cite{Waymo}, PandaSet~\cite{Pandaset}, and KITTI dataset~\cite{KITTI}.
Besides, the individual components of our design are also verified by extensive experiments on WOD~\cite{Waymo}. 
The main contributions of this work include:

\begin{itemize}
\item We propose a Commonsense Prototype-based Detector (\textbf{CPD}) for unsupervised 3D object detection. CPD outperforms state-of-the-art unsupervised 3D detectors by a large margin. 

\item We propose Multi-Frame Clustering (\textbf{MFC}) and CProto-constrained Box Regularization (\textbf{CBR}) for pseudo-label generation and refinement, greatly improving the recall and precision of pseudo-label.

\item We propose CProto-constrained Self-Training (\textbf{CST}) for unsupervised 3D detection. It improves the recognition and localization accuracy of sparse objects, boosting the detection performance significantly.

\end{itemize}

\section{Related Work}
\label{sec:formatting}

\textbf{Fully/weakly supervised 3D object detection.}
Recent fully-supervised 3D detectors build single-stage~\cite{Voxelnet, PointPillars, CenterPoint, SA-SSD, SE-SSD, DSVT}, two-stage~\cite{PointRCNN, PartA2, PV-RCNN, Voxel-RCNN, STD, SFD, TED, VirConv} or multiple stage~\cite{CasA, 3dcas} deep networks for 3D object detection. However, these methods heavily rely on a large amount of precise annotations. Some weakly supervised methods replace the box annotation with low-cost click annotation\cite{TowardsFramework}.  Other methods decrease the supervision by only annotating a part of scenes ~\cite{NoiseDet,3DIoUMatch, SESS, ATF-3D} or a part of instances~\cite{CoIn}. Unlike all of the above works, we aim to design a 3D detector that does not require human-level annotations.

\begin{figure*}[t]
  \centering
   \includegraphics[width=1\linewidth]{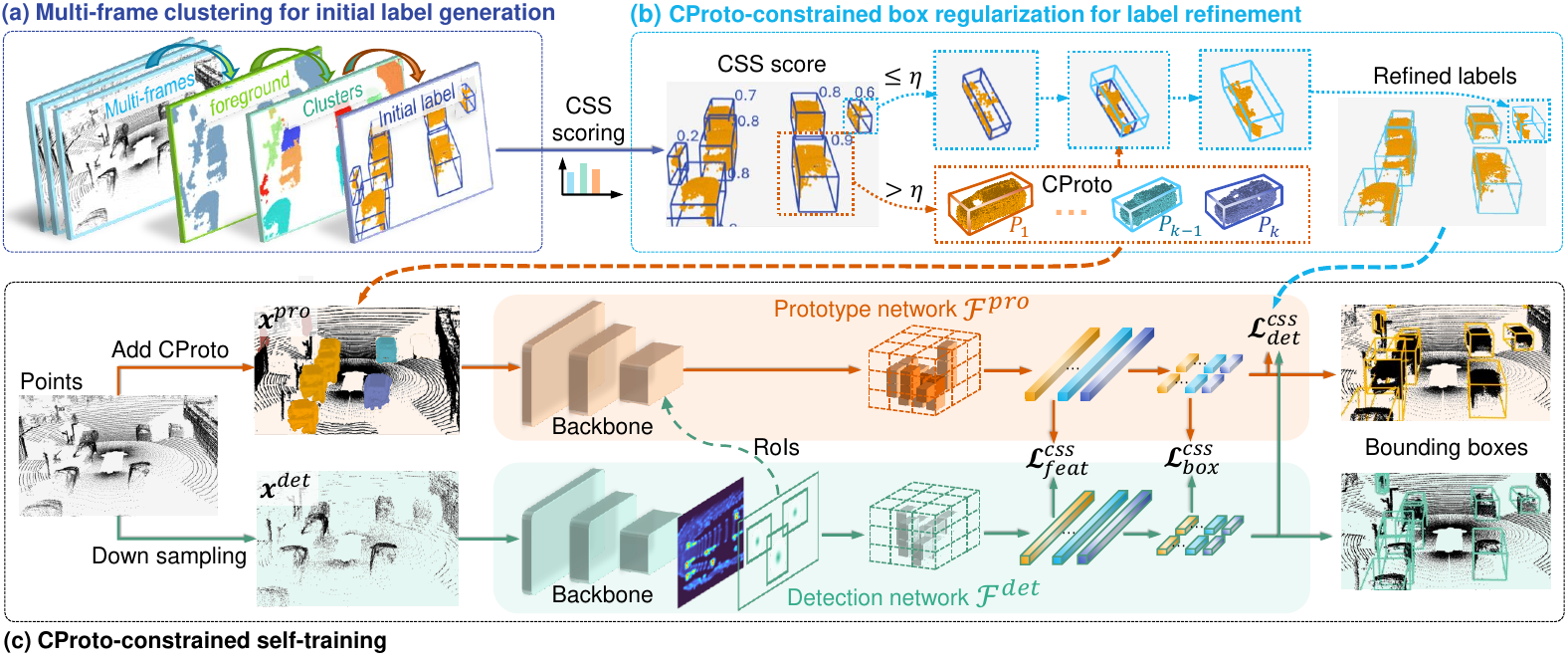}
   \caption{CPD framework. (a) Initial pseudo-labels are generated by multi-frame clustering. (b) The commonsense prototype (CProto) is constructed from high-quality pseudo-labels based on CSS score. The low-quality labels are further refined by the shape prior from CProto. (c) A prototype network fed with dense points from CProto produces high-quality features to guide the detection network convergence.}
   \label{fig:framework}
\end{figure*}

\textbf{Unsupervised 3D object detection.}
Previous unsupervised pre-training methods discern latent patterns within the unlabeled data by masked labels~\cite{GD-MAE} or contrastive loss~\cite{Proposalcontrast,GCC-3D}. But these methods require human labels for fine-turning.
Traditional methods~\cite{hybrid_approach, MODT,datmo} employ ground removal and clustering for 3D object detection without human labels, but suffer from poor detection performance. Some deep learning-based methods generate pseudo-labels by clustering and use the pseudo-labels to train a 3D detector \cite{MODEST} iteratively. Recent OYSTER \cite{OYSTER} improves pseudo-label quality with temporal consistency. However, most pseudo-labels of incomplete objects cannot be recovered by temporal consistency. Our CPD addresses this problem by leveraging the geometry prior from CProto to refine the pseudo-label and guide the network convergence. 

\textbf{Prototype-based methods.}
The prototype-based methods are widely used in 2D detection \cite{FSOD, MoCA, ICPE, rpn_prototype, CPNet} when novel classes are incorporated. Inspired by these methods, 
Prototypical VoteNet \cite{ProtoVote} constructs geometric prototypes learned from basic classes for few-shot 3D object detection.
GPA-3D \cite{GPA-3D} and CL3D \cite{CL3D} build geometric prototypes from a source-domain model for domain adaptive 3D detection. However, both the learning from basic class and training on the source domain require high-quality annotations. 
Unlike that, we construct CProto using commonsense knowledge and detect 3D objects in a zero-shot manner without human-level annotations.
\section{CPD Method}
This paper introduces the Commonsense Prototype-based Detector (CPD), a novel approach for unsupervised 3D object detection. As shown in Fig.~\ref{fig:framework}, CPD consists of three main parts: (1) initial label generation; (2) label refinement; (3) self-training. We detail the designs as follows.

\subsection{Initial Label Generation}

Recent unsupervised methods~\cite{MODEST,OYSTER} detect 3D objects in a class-agnostic way. How to classify objects  (e.g. vehicle and pedestrian) without annotation is still an unsolved challenge.
Our observations indicate that some stationary objects in consecutive frames, appear more complete (see Fig.\ref{fig:motivation} (f)) and can be classified by predefined sizes. 
This motivates us to design a Multi-Frame Clustering (MFC) method to generate initial labels. 
MFC involves motion artifact removal, clustering, and post-processing.

\textbf{Motion Artifact Removal (MAR).} 
Directly transforming and concatenating $2n+1$ consecutive frames $\{\boldsymbol{x}_{-n}, ..., \boldsymbol{x}_{n}\}$ (i.e., past $n$, future $n$, and the current frame) into a single point cloud $\boldsymbol{x}_0^*$ introduces motion artifacts from moving objects, leading to increased label errors as the $n$ grows (see Fig.~\ref{fig:mar}(a)). 
To mitigate this issue, we first transform the consecutive frames to global system and calculate the Persistence Point Score (PPScore)\cite{MODEST} by consecutive frames to identify the points in motion. 
We keep all the points from $\boldsymbol{x}_0$ and remove moving points from the other frames $\boldsymbol{x}_{-n},...,\boldsymbol{x}_{-1},\boldsymbol{x}_{1},...,\boldsymbol{x}_{n}$. 
After this removal, we concatenate the frames to obtain dense points $\boldsymbol{x}_0^*$. 

\textbf{Clustering and post-processing.} 
In line with recent study~\cite{OYSTER}, we apply the ground removal\cite{GroundRemoval}, DBSCAN~\cite{DBSCAN} and bounding box fitting~\cite{BoxFit} on $\boldsymbol{x}_0^*$ to obtain a set of class-agnostic bounding boxes $\hat{\boldsymbol{b}}$. 
We observe that the objects of the same class typically have similar sizes in 3D space. Therefore, we pre-define class-specific size thresholds (e.g. the length of vehicle is generally larger than 0.5m) based on human commonsense to classify  $\hat{\boldsymbol{b}}$ into different categories. 
We then apply class-agnostic tracking to associate the small background objects with foreground trajectories, and enhance the consistency of objects' sizes by using temporal coherency~\cite{OYSTER}. This process results in a set of initial pseudo-labels $\boldsymbol{b}=\{b_j\}_j$, where $b_j=[x,y,z,l,w,h,\alpha,\beta,\tau]$ represents position, width, length, height, azimuth angle, class identity, and tracking identity, respectively.

\subsection{CProto-constrained Box Regularization for Label Refinement}

\begin{figure}[t]
  \centering
   \includegraphics[width=1\linewidth]{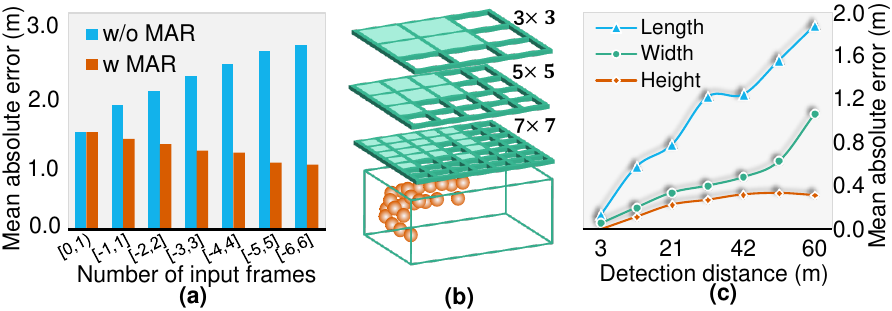}
   \caption{(a) Length absolute error with different frames. (b) Multi-level occupancy score. (c) Mean size error of initial labels.}
   \label{fig:mar}
\end{figure}
As noted in Section~\ref{sec:intro}, initial labels for incomplete objects often suffer from inaccuracies in sizes and positions. 
To tackle this issue, we introduce the CProto-constrained Box Regularization (CBR) method. The key idea is to construct a high-quality CProto set based on unsupervised scoring from complete objects to refine the pseudo-labels of incomplete objects. 
Different from OYSTER~\cite{OYSTER}, which can only refine the pseudo-labels of objects having at least one complete scan, our CBR can refine pseudo-labels of all objects, significantly decreasing the overall size and position errors. 

\begin{figure}[t]
  \centering
   \includegraphics[width=1\linewidth]{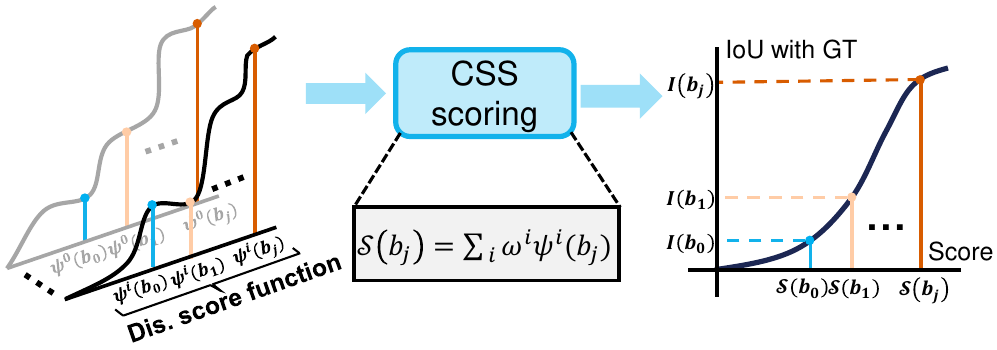}
   \caption{Completeness and size similarity scoring.}
   \label{fig:css_ill}
\end{figure}

\textbf{Completeness and Size Similarity (CSS) scoring.}
Existing label scoring methods such as IoU scoring~\cite{PV-RCNN} are designed for fully supervised detectors.
In contrast, we introduce an \textbf{\textit{unsupervised}} Completeness and Size Similarity scoring (CSS) method. It aims to approximate the IoU score using commonsense knowledge alone (see Fig.~\ref{fig:css_ill}). 

\textit{Distance score.} CSS first assesses the object completeness based on distance, assuming labels closer to the ego vehicle are likely to be more accurate. 
For an initial label $b_j$, we normalize the distance to the ego vehicle within the range [0,1] to compute the distance score as 
\begin{equation}
    \psi^1(b_j)=1-\mathcal{N}(\Vert c_j\Vert),
\end{equation}
where $\mathcal{N}$ is the normalization function and $c_j$ is the location of $b_j$. 
However, this distance-based approach has its limitations. For example, occluded objects near the ego vehicle, which should receive lower scores, are inadvertently assigned high scores due to their proximity. 
To mitigate this issue, we introduce a Multi-Level Occupancy (MLO) score, further detailed in Fig.~\ref{fig:mar} (b).

\textit{MLO score.}
Considering the diverse sizes of objects, we divide the bounding box of the initial label into multiple grids with different length and width resolutions. The MLO score is then calculated by determining the proportion of grids occupied by cluster points, via
\begin{equation}
    \psi^2(b_j)=\frac{1}{N^o}\sum\nolimits_k \frac{O^k}{(r^k)^2},
\end{equation}
where $N^o$ denotes resolution number, $O^k$ is the number of occupied grids under $k$-th resolution, and $r^k$ is the grid number of $k$-th resolution. 

\textit{Size Similarity (SS) score.} 
While the distance and MLO scores effectively evaluate the localization and size quality, they fall short in assessing classification quality. 
To bridge this gap, we introduce the SS score. This score utilizes a class-specific template box $a$ (average size of typical objects in Wikipedia) and calculates a truncated KL divergence~\cite{kl}. Note that, this score is decided by ratio difference, rather than their specific values. Simple commonsense of $l,w,h$ ratios (2:1:1 for Vehicle, 1:1:2 for Pedestrian, 2:1:2 for Cyclist) can also be used here.
\begin{equation}
\psi^3(b_j)=1-min(0.05,\sum\nolimits_\sigma q^b_\sigma log(\frac{q^b_\sigma}{q^a_\sigma}))/0.05,
\end{equation}
where $q_\sigma^a\in\{l^a,w^a,h^a\}, q_\sigma^b\in\{l^b,w^b,h^b\}$ refer to the normalized length, width, and height of the template and label.

We linearly combine the three metrics $\mathcal{S}(b_j)=\sum\nolimits_i \omega^i \psi^i(b_j)$ to produce final scoring, where $\omega^i$ is the weighting factor (in this study we adopt a simple average, $\omega^i=1/3$). 
For each $b_j\in\boldsymbol{b}$, we compute its CSS score $s^{css}_j = \mathcal{S}(b_j)$ and obtain a set of scores $\boldsymbol{s}=\{s_j^{css}\}_j$.

\textbf{CProto set construction.}
Regular learnable prototype-based methods require annotations~\cite{ProtoVote,GPA-3D}, which are unavailable in the unsupervised problem. 
We construct a high-quality CProto set $\boldsymbol{P}=\{P_k\}_k$, representing geometry and size centers based on the \textit{unsupervised} CSS score.
Here, $P_k=\{x^p_k,b^p_k\}$, where $x^p_k$ indicates the inside points, and $b^p_k$ refers to the bounding box. 
Specifically, we first categorize the initial labels $\boldsymbol{b}$ into different groups based on their tracking identity $\tau$. 
Within each group, we select the high-quality boxes and inside points that meet a high CSS score threshold $\eta$ (determined on validation set, using 0.8 in this study). 
Then, we transform all points and boxes into a local coordinate system, and obtain $b^p_k$ by averaging the high-quality boxes and $x^p_k$ by concatenating all the points.

\begin{figure}[t]
  \centering
\includegraphics[width=1\linewidth]{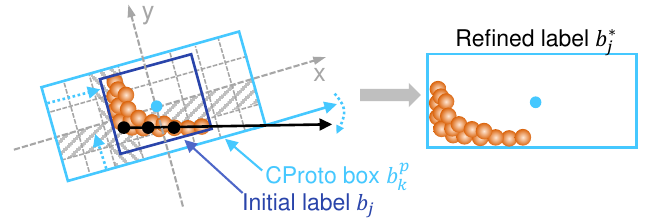}
   \caption{The size of the initial label is replaced by the CProto box, and the position is also corrected.}
   \label{fig:cbr}
\end{figure}

\textbf{Box regularization.}
We next regularize the initial labels by the size prior from CProto. Based on the statistics on WOD validation set~\cite{Waymo}, we observe that the height of the initial labels is relatively correct than length and width (see Fig.~\ref{fig:mar} (c)). Intuitively, the intra-class 3D objects with the same height have similar length and width. Therefore, we associate the initial label $b_j$ with  CProto $P_k$ by the minimum difference in box height. The initial pseudo-labels with the same $P_k$ and similar length and width are naturally classified into the same group.
We then perform re-size and re-localization for each group to refine the pseudo-labels.
(1) \textit{Re-size}. We directly replace the size of $b_j$ using the length, width, and height of $b_k^p \in P_k$. 
(2) \textit{Re-location}.  
Since points are mostly on the object's surface and boundary, we divide the object into different bins and align the box boundary and orientation to the boundary point of the densest part (see Fig.~\ref{fig:cbr}). 
Finally, we obtain improved pseudo-labels $\boldsymbol{b}^*=\{b^*_j\}_j$.

\subsection{CProto-constrained Self-Training (CST)}
Recent methods~\cite{MODEST,OYSTER} utilize pseudo-labels for training 3D detectors. 
However, even after refinement, 
some pseudo-labels remain inaccurate, diminishing the effectiveness of correct supervision and potentially misleading the training process.
To tackle these issues, we propose two designs:
(1) \textit{CSS-Weighted Detection Loss}, which assigns different training weights based on label quality to suppress false supervision signals.
(2) \textit{Geometry Contrast Loss}, which aligns predictions of sparsely scanned points with the dense CProto, thereby improving feature consistency.

\textbf{Network architecture.}
We adopt a dense-sparse alignment architecture (Fig.~\ref{fig:framework} (c)), consisting of a prototype network $\mathcal{F}^{pro}$ and a detection network $\mathcal{F}^{det}$, constructed from two-stage CenterPoint~\cite{CenterPoint}. 
During training, for each $b^*_j$, we add its corresponding points $x^p_k$ from CProto $P_k$ to the scene to obtain a dense point cloud $\boldsymbol{x}^{pro}$.
We feed $\boldsymbol{x}^{pro}$ to $\mathcal{F}^{pro}$ to produce relatively good features and detections. We then feed randomly downsampled points $\boldsymbol{x}^{det}$ as a sparse sample to the $\mathcal{F}^{det}$. We align the features and detections from two branches by the detection loss and contrast loss. During testing, we feed points without downsampling to the detection network $\mathcal{F}^{det}$ to perform detection.

\textbf{CSS weight.}
Considering that the false pseudo-labels may mislead the network convergence, we first calculate a loss weight based on different label qualities.
Formally, we convert a CSS score $s^{css}_i$ of a pseudo-label to 
\begin{align}
\omega_i = \left \{  
\begin{array}{ll}
   0  &  s^{css}_i<S_L\\
   \frac{s^{css}_i-S_L}{S_H-S_L}  & S_L<s^{css}_i<S_H \\
   1  & s^{css}_i>S_H
\end{array},\right.
\end{align}
where $S_H$ and $S_L$ are high/low-quality thresholds (we empirically set 0.7 and 0.4, respectively).

\textbf{CSS-weighted detection loss.} 
To decrease the influence of false labels, we formulate the CSS-weighted detection loss to refine $N$ proposals 
\begin{equation}
    \mathcal{L}_{det}^{css}= \frac{1}{N}\sum \nolimits_i\omega_i(\mathcal{L}^{pro}_i+\mathcal{L}^{det}_i),
\end{equation}
where $\mathcal{L}^{pro}_i$ and $\mathcal{L}^{det}_i$ are detection losses~\cite{Voxel-RCNN} of $\mathcal{F}^{pro}$ and $\mathcal{F}^{det}$, respectively. The losses are calculated by pseudo-labels $\boldsymbol{b}^*$ and network predictions.

\textbf{Geometry contrast loss.}
We formulate two contrast losses that minimize the feature and predicted box difference between the prototype and detection network.
\textit{ (1) Feature contrast loss.}
For a foreground RoI $r_i$ from the detection network, we extract features $\boldsymbol{f}^p_i$ from the prototype network by voxel set abstract~\cite{Voxel-RCNN}, and extract features $\boldsymbol{f}^d_i$ from detection network. We then formulate the contrast loss by cosine distance: 
\begin{equation}
    \mathcal{L}^{css}_{feat}=-\frac{1}{N^f}\sum \nolimits_i \omega_i \frac{\boldsymbol{f}^d_i\cdot\boldsymbol{f}^p_i}{\Vert \boldsymbol{f}^d_i\Vert\Vert \boldsymbol{f}^p_i\Vert},
\end{equation}
where $N^f$ is the foreground proposal number.
\textit{ (2) Box contrast loss.}
For a box prediction $d^p_i$ from the prototype network and a box prediction $d^d_i$ from the detection network. We then formulate the box contrast loss by IoU, location difference, and angle difference: 
\begin{align}
\mathcal{L}^{css}_{box}= &
       \frac{1}{N^f}\sum \nolimits_i \omega_i [1-I(d^d_i,d^p_i) \notag\\
      &+\Vert c^d_i-c^p_i\Vert + |sin(\alpha^d_i-\alpha^p_i)|],
\end{align}
where $I$ denote IoU function; $c^d_i,\alpha^d_i$ refers to position and angle of $d^d_i$; $c^p_i,\alpha^p_i$ refers to position and angle of $d^p_i$. 
We finally summat all losses to training the detector.

\begin{table*}[htbp]
  \centering
      \resizebox{\textwidth}{!}{
    \begin{tabular}{l | c c  c c|c c c c|c c  c c}
    \hline
    \multirow{3}*{Method} & \multicolumn{4}{c|}{Vehicle 3D AP} & \multicolumn{4}{c|}{Pedestrian 3D AP} & \multicolumn{4}{c}{Cyclist 3D AP} \\
    
                            & \multicolumn{2}{c}{L1} & \multicolumn{2}{c|}{L2} & \multicolumn{2}{c}{L1} & \multicolumn{2}{c|}{L2} & \multicolumn{2}{c}{L1} & \multicolumn{2}{c}{L2} \\
    
                            & $IoU_{0.5}$ & $IoU_{0.7}$ & $IoU_{0.5}$ & $IoU_{0.7}$& $IoU_{0.3}$ & $IoU_{0.5}$& $IoU_{0.3}$ & $IoU_{0.5}$& $IoU_{0.3}$ & $IoU_{0.5}$& $IoU_{0.3}$ & $IoU_{0.5}$\\
    \hline
    DBSCAN~\cite{DBSCAN} & 2.32  & 0.29  & 1.94  & 0.25  & 0.51  & 0.00     & 0.19  & 0.00     & 0.28  & 0.03  & 0.20   & 0.00 \\

    DBSCAN init-train~\cite{MODEST} & 17.36 & 2.65  & 14.87 & 2.29  & 1.65  & 0.00     & 1.35  & 0.00     & 0.48  & 0.25  & 0.43  & 0.20 \\

    MODEST~\cite{MODEST} & 18.51 & 6.46  & 15.83 & 5.48  & 11.83 & 0.17  & 8.96  & 0.10   & 1.47  & 1.14  & 1.17  & 1.01 \\

    OYSTER~\cite{OYSTER} & 30.48 & 14.66 & 26.21 & 14.10  & 4.33  & 0.18  & 3.52  & 0.14  & 1.27  & 0.33  & 1.24  & 0.32 \\


    Proto-vanilla & 35.22 & 20.19 & 31.58 & 18.36 & 17.60  & 10.34 & 14.62 & 8.59  & 4.21  & 3.45  & 3.80   & 3.31 \\
    \rowcolor{light} \textbf{CPD (Ours)} & \textbf{57.79} & \textbf{37.40} & \textbf{50.18} & \textbf{32.13} & \textbf{21.91} & \textbf{16.31} & \textbf{18.01} & \textbf{13.22} & \textbf{5.83} & \textbf{5.06} & \textbf{5.61} & \textbf{4.87} \\
    \hline
    \end{tabular}%
    }
        \caption{Unsupervised 3D object detection results on WOD validation set. The results of previous methods are reproduced by us. }
  \label{tab:unsup-waymo}%
\end{table*}%
\section{Experiments}

\subsection{Datasets}
\textbf{Waymo Open Dataset (WOD).}
We conducted extensive experiments on the WOD~\cite{Waymo} due to its diverse scenes. The WOD contains 798, 202 and 150 sequences for training, validation and testing, respectively. We adopted similar metrics (3D AP L1 and L2) as fully/weakly supervised methods~\cite{TED, CoIn}. \textbf{\textit{No annotations}} were used for training. 

\begin{table}[htbp]
  \centering
    \setlength{\tabcolsep}{6pt}
    \resizebox{\columnwidth}{!}{
    \begin{tabular}{ l | c c c | c c c}
    \hline
    \multirow{2}*{Method} & \multicolumn{3}{c|}{3D AP L1 ($IoU_{0.7,0.5,0.5}$) } & \multicolumn{3}{c}{3D AP L2 ($IoU_{0.7,0.5,0.5}$)} \\
     &  Vehicle & Ped. &Cyclist & Vehicle & Ped. &Cyclist \\
    \hline
    MODEST~\cite{MODEST} & 7.5    & 0.0   & 0.0   & 6.5  & 0.0  & 0.0 \\
    OYSTER~\cite{OYSTER} & 21.6  & 0.6  & 0.0  & 18.7  & 0.5  & 0.0 \\
    \rowcolor{light}\textbf{CPD (Ours)} & \textbf{37.2} & \textbf{18.6} & \textbf{5.7} & \textbf{32.4} & \textbf{16.5} & \textbf{5.5} \\
    \hline
    \end{tabular}%
    }
      \caption{Unsupervised 3D detection results on WOD test set. }
  \label{tab:unsup-wod-test}%
\end{table}%

\begin{table}[htbp]
  \centering
    \setlength{\tabcolsep}{3.5pt}
    \resizebox{\columnwidth}{!}{
    \begin{tabular}{ l | c c c | c c c}
    \hline
    \multirow{2}*{Method} & \multicolumn{3}{c|}{BEV AP } & \multicolumn{3}{c}{BEV Recall} \\
     &  $IoU_{0.3}$ & $IoU_{0.5}$ &$IoU_{0.7}$ & $IoU_{0.3}$ & $IoU_{0.5}$ & $IoU_{0.7}$ \\
    \hline

    MODEST~\cite{MODEST} & 22.0    & 7.5   & 2.8   & 49.7  & 28.9  & 14.9 \\

    OYSTER~\cite{OYSTER} & 43.5  & 29.5  & 18.1  & 62.8  & 44.8  & 28.1 \\
    \rowcolor{light}\textbf{CPD (Ours)} & \textbf{50.7} & \textbf{41.0} & \textbf{24.6} & \textbf{63.1} & \textbf{54.8} & \textbf{37.4} \\
    \hline
    \end{tabular}%
    }
      \caption{The class-agnostic comparison results on the PandaSet dataset, evaluated on the 0-80m detection range.}
  \label{tab:unsup-panda}%
\end{table}%

\textbf{PandaSet dataset.}
To compare with recent unsupervised methods~\cite{OYSTER}, we also conducted experiments on the PandaSet~\cite{Pandaset}. 
Like \cite{OYSTER}, we split the dataset into 73 training and 30 validation snippets and use class-agnostic BEV AP and recall metrics with 0.3, 0.5, and 0.7 IoU thresholds. 

\textbf{KITTI dataset.} 
Since the KITTI detection dataset~\cite{KITTI} did not provide consecutive frames, we only tested our method on the 3769 val split~\cite{Voxel-RCNN}. 
We used similar metrics (Car 3D AP R40 with 0.5 and 0.7 IoU thresholds) as employed in fully/weakly supervised methods~\cite{VirConv, CoIn}.

\subsection{Implementation Details}
\textbf{Network details.}
Both prototype and detection networks adopt the same 3D backbone as CenterPoint~\cite{CenterPoint} and the same RoI refinement network as Voxel-RCNN~\cite{Voxel-RCNN}.
For the WOD and KITTI datasets, we use the same detection range and voxel size as CenterPoint~\cite{CenterPoint}. For the Pandaset, we use the same detection range as OYSTER~\cite{OYSTER}.

\textbf{Training details.}
We adopt the widely used global scaling and rotation data augmentation. 
We trained our network on 8 Tesla V100 GPUs with the ADAM optimizer. We used a learning rate of 0.003 with a one-cycle learning rate strategy. We trained the CPD for 20 epochs.

\subsection{Comparison with Unsupervised Detectors}

\textbf{Results on WOD.}
The results on the WOD validation set and test set are presented in Table~\ref{tab:unsup-waymo} and Table~\ref{tab:unsup-wod-test}. 
All methods use identical size thresholds to define the object classes and use single traversal. 
Our method significantly outperforms existing unsupervised methods. Notably, under the 3D AP L2 with IoU thresholds of 0.7, 0.5, and 0.5, our CPD outperforms OYSTER~\cite{OYSTER} by 18.03\%, 13.08\%, and 4.55\% on Vehicle, Pedestrian, and Cyclist, respectively. 
These advancements come from our MFC, CBR, and CST designs, which yield superior pseudo-labels and enhanced detection accuracy. 
CPD also surpasses the Proto-vanilla method, which uses class-specific prototype~\cite{ProtoNIPS}. 

\begin{table}[htbp]
  \centering
    \setlength{\tabcolsep}{2.pt}
    \resizebox{\columnwidth}{!}{
    \begin{tabular}{ l |c| c c c | c c c}
    \hline
    \multirow{2}*{Method}&\multirow{2}*{Labels} & \multicolumn{3}{c|}{3D AP @ $IoU_{0.5}$ } & \multicolumn{3}{c}{3D AP @ $IoU_{0.7}$} \\
     & & Easy & Mod. & Hard &Easy & Mod. & Hard \\
    \hline
    \textcolor[rgb]{ .192,  .282,  .624}{\textit{CenterPoint~\cite{CenterPoint}}} & \textcolor[rgb]{ .192,  .282,  .624}{\textit{100\%}} &
    \textcolor[rgb]{ .192,  .282,  .624}{\textit{97.07}} & \textcolor[rgb]{ .192,  .282,  .624}{\textit{89.23}}
    &\textcolor[rgb]{ .192,  .282,  .624}{\textit{81.81}} & \textcolor[rgb]{ .192,  .282,  .624}{\textit{88.55}}
    &\textcolor[rgb]{ .192,  .282,  .624}{\textit{78.38}} &\textcolor[rgb]{ .192,  .282,  .624}{\textit{71.43}} \\
    \textcolor[rgb]{ .192,  .282,  .624}{\textit{Sparsely-sup.~\cite{CoIn}}} & \textcolor[rgb]{ .192,  .282,  .624}{\textit{2\%}}
    &\textcolor[rgb]{ .192,  .282,  .624}{\textit{-}} &\textcolor[rgb]{ .192,  .282,  .624}{\textit{-}}
    & \textcolor[rgb]{ .192,  .282,  .624}{\textit{-}} & \textcolor[rgb]{ .192,  .282,  .624}{\textit{49.69}}
    &\textcolor[rgb]{ .192,  .282,  .624}{\textit{31.55}} & \textcolor[rgb]{ .192,  .282,  .624}{\textit{25.91}} \\
    \hline 
    MODEST~\cite{MODEST} & 0     & 47.56 & 33.43 & 30.57 & 12.65 & 11.14 & 10.60 \\
    OYSTER~\cite{OYSTER} & 0     & 65.33 & 54.82 & 43.59 & 23.22 & 20.31 & 19.97 \\
    \rowcolor{light} \textbf{CPD (Ours)} & 0     & \textbf{90.85} & \textbf{81.01} & \textbf{79.80} & \textbf{72.98} & \textbf{55.07} & \textbf{53.94} \\
    \hline
    \end{tabular}%
    }
      \caption{Car detection comparison with fully/weakly supervised detectors on KITTI val set. The models are trained on WOD.}
  \label{tab:sup-kitti}%
\end{table}%

\begin{table}[htbp]
  \centering
    \setlength{\tabcolsep}{5.pt}
    \resizebox{\columnwidth}{!}{
    \begin{tabular}{ l |c| c c | c c}
    \hline
    \multirow{2}*{Method}&\multirow{2}*{Labels} & \multicolumn{2}{c|}{3D AP L1 }&  \multicolumn{2}{c}{3D AP L2 } \\
     &  & $IoU_{0.5}$ & $IoU_{0.7}$ & $IoU_{0.5}$ & $IoU_{0.7}$ \\
    \hline
    \textcolor[rgb]{ .192,  .282,  .624}{\textit{CenterPoint~\cite{CenterPoint}}} & \textcolor[rgb]{ .192,  .282,  .624}{\textit{100\%}} &
    \textcolor[rgb]{ .192,  .282,  .624}{\textit{89.23}} & \textcolor[rgb]{ .192,  .282,  .624}{\textit{73.72}}
    &\textcolor[rgb]{ .192,  .282,  .624}{\textit{78.52}} & \textcolor[rgb]{ .192,  .282,  .624}{\textit{65.52}} \\
    \textcolor[rgb]{ .192,  .282,  .624}{\textit{Sparsely-sup.~\cite{CoIn}}} & \textcolor[rgb]{ .192,  .282,  .624}{\textit{2\%}}
    &\textcolor[rgb]{ .192,  .282,  .624}{\textit{-}} &\textcolor[rgb]{ .192,  .282,  .624}{\textit{32.15}}
    & \textcolor[rgb]{ .192,  .282,  .624}{\textit{-}} & \textcolor[rgb]{ .192,  .282,  .624}{\textit{27.97}}\\
    \hline
    MODEST~\cite{MODEST} & 0     & 18.51 & 6.46  & 15.83 & 5.48 \\
    OYSTER~\cite{OYSTER} & 0     & 30.48 & 14.66 & 26.21 & 14.60 \\
     \rowcolor{light} \textbf{CPD (Ours)} & 0 &  \textbf{57.79} & \textbf{37.40} & \textbf{50.18} & \textbf{32.13} \\
    \hline
    \end{tabular}%
    }
      \caption{Vehicle detection comparison with fully/weakly supervised detectors on WOD validation set.}
  \label{tab:sup-waymo}%
\end{table}%

\textbf{Results on PandaSet.}
The class-agnostic results on PandaSet are presented in Table~\ref{tab:unsup-panda}.
Our method outperforms OYSTER by 6.5\% AP and 9.3\% Recall under 0.7 IoU threshold. 
This improvement is largely due to our CPD's enhanced label quality. 
Unlike OYSTER, which suffers from the misleading effects of false labels during training, our CPD leverages the size prior from CProto to significantly improve these labels.

\subsection{Comparison with Fully/Weakly Supervised Detectors}

\textbf{Results on KITTI dataset.}
To further validate our method, we pre-trained our CPD, along with OYSTER~\cite{OYSTER} and MODEST~\cite{MODEST}, on WOD and tested them on the KITTI dataset using Statistical Normalization (SN)~\cite{SN}. 
The car detection results are in Table~\ref{tab:sup-kitti}. 
We first compared our method with a sparsely supervised method (weakly supervised with 2\% labels)~\cite{CoIn} that annotates a single instance per frame for training. 
Our unsupervised CPD outperforms this sparsely supervised method by 23.52\% 3D AP @ $IoU_{0.7}$ on moderate car class. 
Additionally, our method attains 90.85\% and 81.01\% 3D AP for the easy and moderate car classes at a 0.5 IoU threshold. 
Notably, this performance is comparable to that of the fully supervised method CenterPoint~\cite{CenterPoint}, demonstrating the advancement of our method.

\begin{figure}[t]
  \centering
   \includegraphics[width=1\linewidth]{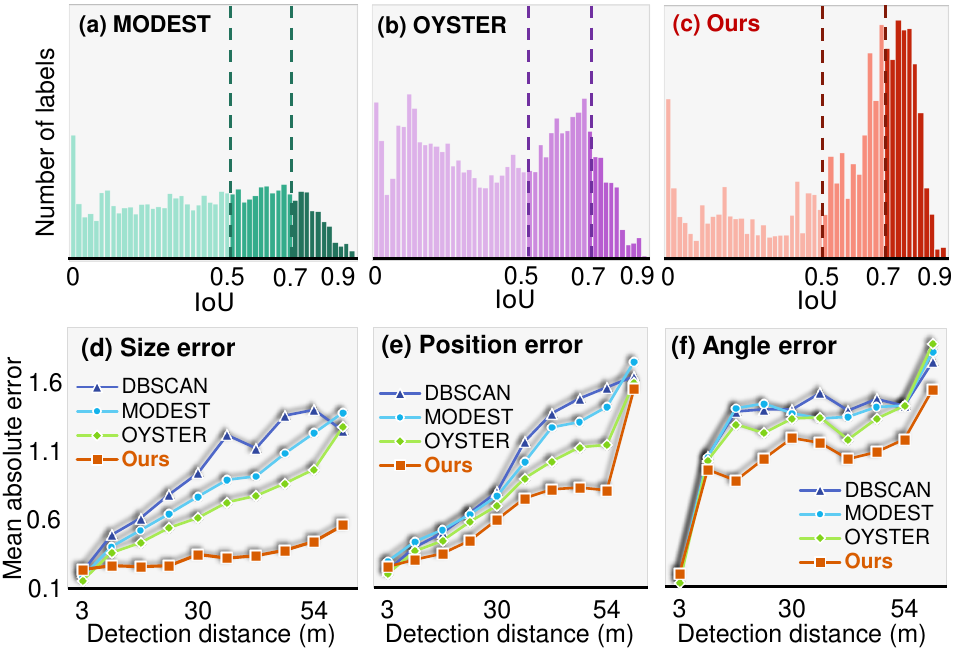}
   \caption{(a-c) IoU distribution between pseudo-labels and ground truth. (d-f) Mean absolute error associated with the size, position, and angle of pseudo-labels generated by different methods.  }
   \label{fig:exp_label_compare}
\end{figure}
\begin{table}[htbp]
  \centering
    \setlength{\tabcolsep}{4.pt}
    \resizebox{\columnwidth}{!}{
    \begin{tabular}{ l | c c c | c c c}
    \hline
    \multirow{2}[4]{*}{Method} & \multicolumn{3}{c|}{3D Recall} & \multicolumn{3}{c}{3D Precision} \\
    & $IoU_{0.3}$ & $IoU_{0.5}$ &$IoU_{0.7}$ &$IoU_{0.3}$ & $IoU_{0.5}$ & $IoU_{0.7}$\\
    \hline
    DBSCAN~\cite{DBSCAN} & 22.85 & 16.44 & 6.52  & 29.41 & 21.16 & 8.39 \\
    MODEST~\cite{MODEST} & 17.35 & 12.04 & 4.89  & 32.28 & 22.81 & 10.05 \\
    OYSTER~\cite{OYSTER} & 31.10  & 21.01 & 11.12 & 31.22 & 21.09 & 9.45 \\
    \rowcolor{light}\textbf{Ours} & \textbf{45.66} & \textbf{39.33} & \textbf{20.54} & \textbf{34.17} & \textbf{28.22} & \textbf{14.74} \\
    \hline
    \end{tabular}%
    }
      \caption{Pseudo-label comparison results on WOD validation set.}
  \label{tab:label-waymo}%
\end{table}%

\textbf{Results on WOD.}
We also compared our method with fully/weakly supervised methods on the WOD validation set~\cite{Waymo}. The vehicle detection results are in Table~\ref{tab:sup-waymo}. Our unsupervised CPD outperforms the sparsely supervised method (2\% annotation) by 5.25\% and 4.16\% in terms of 3D AP L1 and L2 respectively. 

\subsection{Pseudo-label Comparison} 
To validate our pseudo-labels, we analyzed their 3D recall and precision on the WOD validation set. 
As shown in Table~\ref{tab:label-waymo}, our method surpasses the previous best-performing OYSTER with a 9.42\% recall and 5.29\% precision improvement (under a 0.7 IoU threshold). 
To understand the sources of this improvement, we examined the IoU between the pseudo-labels and ground truth, and compared the IoU distributions in Fig.~\ref{fig:exp_label_compare} (a)(b)(c). 
We also present the mean absolute error of size, position, and angle between different pseudo-labels in Fig.~\ref{fig:exp_label_compare} (d)(e)(f). The IoU distribution of our method is much closer to 1 than other methods, 
and it also exhibits lower errors in size, position, and angle. These results verify that our MFC and CBR significantly reduce label errors. 

\begin{figure}[t]
  \centering
   \includegraphics[width=1\linewidth]{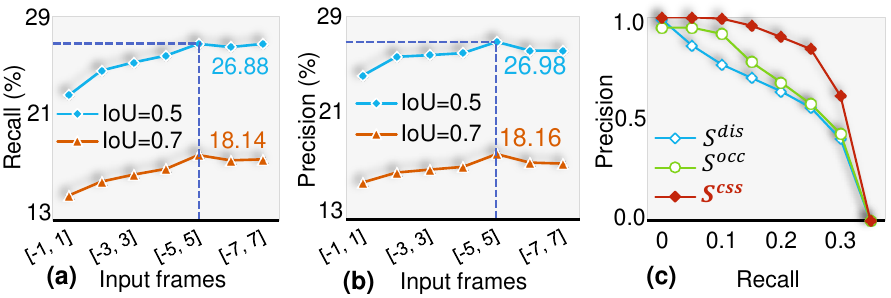}
   \caption{(a)(b)The recall and precision of initial pseudo-labels by using different frames. (c) The recall-precision curve of initial pseudo-labels by using different scores. }
   \label{fig:exp_frames}
\end{figure}

\begin{table}[htbp]
  \centering
      \setlength{\tabcolsep}{4.pt}
    \resizebox{\columnwidth}{!}{
    \begin{tabular}{c c c c  | c  c | c c}
    \hline
    \multicolumn{4}{c|}{Components} & \multicolumn{2}{c|}{3D AP L1} & \multicolumn{2}{c}{3D AP L2} \\

    SFC & MFC  & CBR   & CST     & $IoU_{0.5}$ &$IoU_{0.7}$ & $IoU_{0.5}$ &$IoU_{0.7}$ \\
    \hline
    $\checkmark$ &  &  &    & 17.36 & 2.65& 14.87 & 2.29 \\

          & $\checkmark$     &  &  &19.91 & 5.01& 18.31 & 4.77 \\

          & $\checkmark$     & $\checkmark$     &   & 48.26 & 28.01& 41.69 & 24.04 \\

          & $\checkmark$     & $\checkmark$     & $\checkmark$     &  \textbf{57.79} & \textbf{37.40} & \textbf{50.18} & \textbf{32.13}\\
    \hline
    \end{tabular}%
    }
      \caption{ CPD component analysis results on WOD validation set.}
  \label{tab:ablation-cpd}%
\end{table}%
\begin{figure*}[t]
  \centering
   \includegraphics[width=1\linewidth]{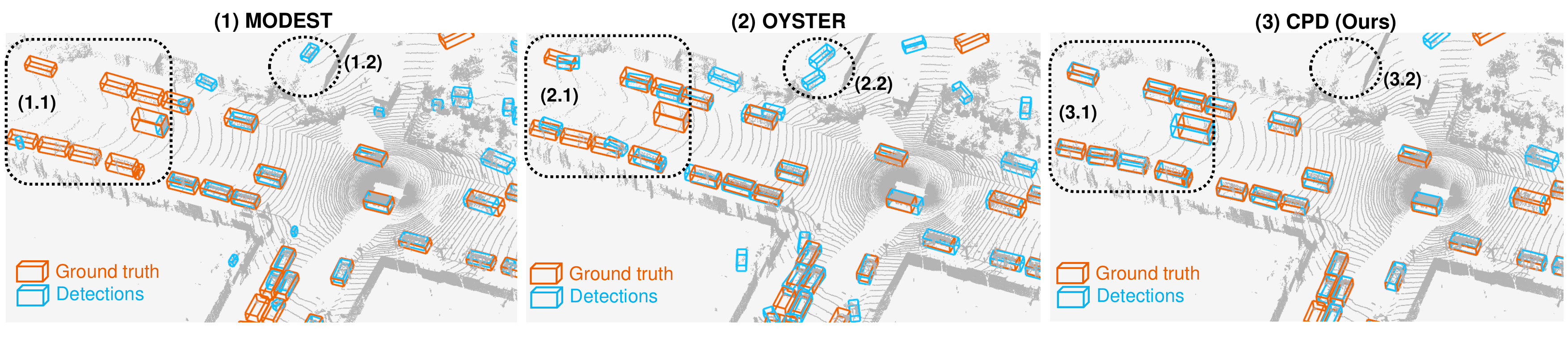}
   \caption{Visualization comparison of different detection results on WOD validation set.}
   \label{fig:vi}
\end{figure*}
\begin{table}[htbp]
  \centering
      \small
    \begin{tabular}{l |l l l}
    \hline
    \multirow{2}*{CSS Components} & \multicolumn{3}{c}{ BEV AP}\\
 & $IoU_{0.3}$ & $IoU_{0.5}$ & $IoU_{0.7}$ \\
    \hline
    Distance & 32.70 & 24.57 & 14.80 \\

    Distance+MLO & 34.40 & 26.25 & 15.91 \\

    Distance+MLO+SS & \textbf{38.95} & \textbf{31.06} & \textbf{19.49} \\
    \hline
    \end{tabular}%
    
      \caption{CSS component analysis results on WOD validation set.}
  \label{tab:ablation-css}%
\end{table}%

\subsection{Ablation Study}
\textbf{Components analysis of CPD.}
To evaluate the individual contributions of our designs, we incrementally added each component and assessed their impact on vehicle detection using the WOD validation set.
The results are shown in Table~\ref{tab:ablation-cpd}. 
Our MFC method surpasses Single Frame Clustering (SFC) by 2.52\% in AP, attributed to the more complete point representation of objects across consecutive frames compared to a single frame.
The CBR further enhances performance by 19.27\% in AP, as it reduces size and location errors in pseudo-labels. 
The CST contributes an 8.09\% increase in AP, demonstrating the effectiveness of geometric features from CProto in detecting sparse objects.

\textbf{Frame number of MFC.}
To examine the effect of frame count on initial pseudo-label quality, we experimented with different numbers of past and future point cloud frames on the WOD validation set. The BEV results, shown in Fig.~\ref{fig:exp_frames} (a)(b), indicate optimal performance with [-5, 5] frames (five past, five future, and the current frame). Additional frames did not significantly improve recall or precision. Consequently, we used 11 frames for initial pseudo-label generation in this study. 

\begin{table}[htbp]
  \centering
      \setlength{\tabcolsep}{4.pt}
    \resizebox{\columnwidth}{!}{
    \begin{tabular}{l |l l|l l}
    \hline
    \multirow{2}*{CBR Components} & \multicolumn{2}{c|}{BEV Recall} & \multicolumn{2}{c}{BEV Precision} \\
 & $IoU_{0.5}$ &$IoU_{0.7}$ & $IoU_{0.5}$ & $IoU_{0.7}$ \\
    \hline
    MFC  & 26.88 & 18.14 & 26.98 & 18.16 \\

    MFC+Re-size & 30.79 & 21.54 & 29.43 & 21.33 \\

    MFC+Re-size+Re-localization & \textbf{43.47} & \textbf{27.97} & \textbf{30.90} & \textbf{21.62} \\
    \hline
    \end{tabular}%
    }
      \caption{CBR component analysis results on WOD validation set.}
  \label{tab:ablation-cbr}%
\end{table}%

\textbf{Component analysis of CSS Scoring.}
To assess the effectiveness of our scoring system, we calculated the BEV AP of initial pseudo-labels with different scores. 
These evaluations, reported in Table~\ref{tab:ablation-css}, show that incorporating all components (distance, MLO, and SS) yields the highest AP. 
The recall-precision curve, plotted in Fig.~\ref{fig:exp_frames} (c), also supports this finding. These indicate the significance of each component in accurately measuring pseudo-label quality. 

\textbf{Components analysis of CBR.}
To evaluate the impact of re-sizing and re-localization in CBR, we conducted experiments and analyzed pseudo-label performance. 
As shown in Table~\ref{tab:ablation-cbr}, 
re-sizing results in a 3.91\% and 3.4\% increase in BEV recall at the 0.5 and 0.7 IoU thresholds, respectively; 
re-localization further enhances recall by 12.68\% and 6.43\% at these thresholds, while also increasing precision. 
These results indicate the importance of both components, which effectively refine pseudo-labels.

\textbf{Components analysis of CST.} 
To assess the effectiveness of each component in CST, we established a baseline using only CBR-generated pseudo-labels for training a two-stage CenterPoint detector, then incrementally added our loss components and evaluated vehicle detection performance on the WOD validation set.
As shown in Table~\ref{tab:ablation-cst}, all loss components contribute to performance improvement. 
Specifically, our $\mathcal{L}^{css}_{det}$ mitigates the influence of false pseudo-label using CSS weight, and improves the 3D AP L2 at $IoU_{0.7}$ by 4.79\%. 
Our $\mathcal{L}^{css}_{feat}$ and $\mathcal{L}^{css}_{box}$ improve the 3D AP L2 at $IoU_{0.7}$ by 0.75\% and 2.55\% respectively, through leveraging geometric knowledge from dense CProto for more effective sparse object detection.

\subsection{Visualization Comparison}

\begin{table}[htbp]
  \centering
      \setlength{\tabcolsep}{5.pt}
    \resizebox{\columnwidth}{!}{
    \begin{tabular}{l |l l|l l}
    \hline
    \multirow{2}*{CST Components} & \multicolumn{2}{c|}{3D AP L1} & \multicolumn{2}{c}{3D AP L2} \\
 & $IoU_{0.5}$ &$IoU_{0.7}$ & $IoU_{0.5}$ & $IoU_{0.7}$ \\
    \hline
    CBR-only & 48.26 & 28.01 & 41.69 & 24.04 \\

    CBR+$\mathcal{L}^{css}_{det}$ & 49.31 & 29.78 & 42.50  & 28.83 \\

    CBR+$\mathcal{L}^{css}_{det}$+$\mathcal{L}^{css}_{feat}$ & 52.01 & 32.17 & 44.12 & 29.58 \\

   CBR+$\mathcal{L}^{css}_{det}$+$\mathcal{L}^{css}_{feat}$+$\mathcal{L}^{css}_{box}$ &  \textbf{57.79} & \textbf{37.40} & \textbf{50.18} & \textbf{32.13} \\
    \hline

    \end{tabular}%
    }
      \caption{CST component analysis results on WOD validation set.}
  \label{tab:ablation-cst}%
\end{table}%

To provide a more intuitive understanding of how our method improves detection performance, we visually compare our results with those of MODEST~\cite{MODEST} and OYSTER~\cite{OYSTER}, as shown in Fig.~\ref{fig:vi}. 
MODEST often misses distant, sparse objects (Fig.~\ref{fig:vi}(1.1)), while OYSTER detects them but inaccurately reports their sizes and positions (Fig.~\ref{fig:vi}(2.1)). 
In contrast, CPD, using our CProto-based design, not only recognizes these objects but also accurately predicts their sizes and positions (Fig.~\ref{fig:vi}(3.1)). 
Furthermore, since our CST reduces the influence of false pseudo-labels, the false positives (Fig.~\ref{fig:vi}(3.2)) are also much fewer than the previous methods (Fig.~\ref{fig:vi}(1.2)(2.2)).

\section{Conclusion}
This paper presents the CPD framework, a novel approach for accurate unsupervised 3D object detection. First, we develop an MFC method to generate initial pseudo-labels. Then, a CProto set is constructed using CSS scoring. Next, we introduce a CBR method to refine these pseudo-labels. Lastly, a CST is designed to enhance detection accuracy for sparse objects. Extensive experiments have verified the effectiveness of our design. Notably, for the first time, our unsupervised CPD method surpasses some weakly supervised methods, demonstrating the advancement of our approach.

\textbf{Limitations.} One notable limitation of our work is the significantly lower Average Precision (AP) for minority classes, such as cyclists (Table~\ref{tab:unsup-waymo}), compared to more prevalent classes like vehicles. This disparity is largely due to the scarce instances of these minority classes within the dataset. Future efforts to collect such objects could be a promising avenue to tackle this issue.

\textbf{Acknowledgements.}
This work was supported in part by the National Natural Science Foundation of China (No.62171393), the Fundamental Research Funds for the Central Universities (No.20720220064).

{
    \small
    \bibliographystyle{ieeenat_fullname}
    \bibliography{main}
}

\clearpage
\setcounter{page}{1}
\maketitlesupplementary
\begin{figure}[t]
  \centering
   \includegraphics[width=1\linewidth]{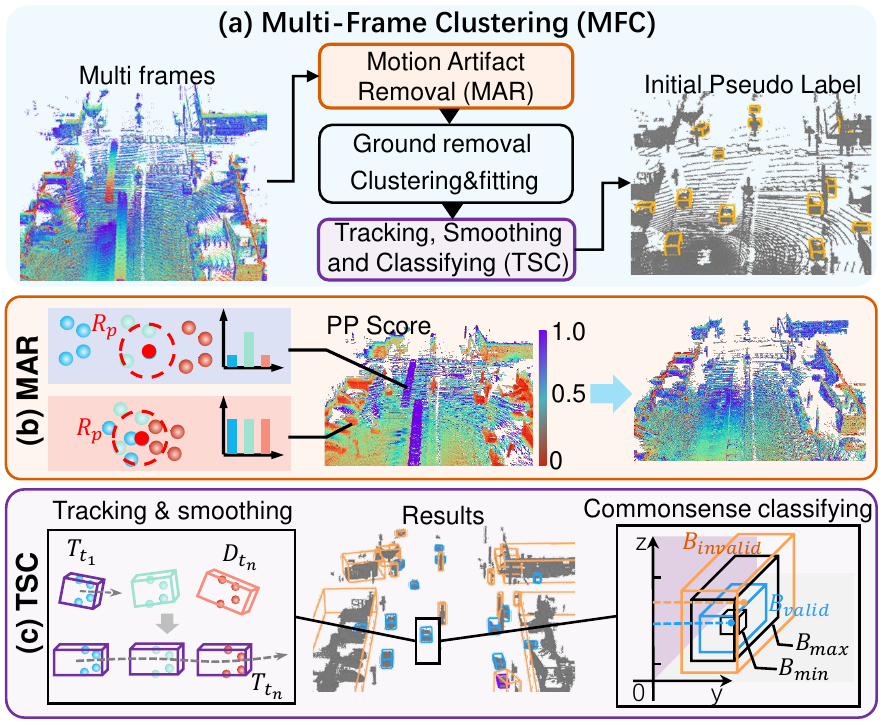}
   \caption{ The MFC consists of motion artifact removal, clustering (ground removal, points clustering, and box fitting), and post-processing ( tracking, smoothing, and classifying).}
   \label{fig:sup-mfc}
\end{figure}

\section{More Details of Method}
\textbf{More details of MFC.}
In our main paper section 3.1, we introduced the Multi-Frame Clustering (MFC) for initial label generation. For a more intuitive understanding, we provide a framework illustration in Fig.~\ref{fig:sup-mfc}. 
Here we present more details of post-processing. As mentioned in our main paper, we pre-defined a set of class-specific size thresholds based on human commonsense to classify pseudo labels into different categories. Taking the WOD  as an example, we pre-define five categories: `Discard Small', `Pedestrian', `Cyclist', `Vehicle', and `Discard Large'.  Formally, for a cluster box $b_j$, we determine the class identity $\beta$ by sequentially matching from the thresholds:

\begin{align}
\footnotesize
\beta = \left \{  
\begin{array}{ll}
   Dis Small  & h\leq0.8,\\
   Vehicle  &  1<h\leq3, 0.5<w\leq3, 0.5<l\leq8,\\
   Pedestrian  & 0.8<h\leq2.3, 0.2<w\leq1., 0.2<l\leq1.,\\
   Cyclist  &  1.4<h\leq2., 0.5<w\leq1., 1.<l\leq2.5,\\
   Dis Large  & others. \\
\end{array}\right.
\end{align}
Where $l,w,h$ refers to the length, width, and height of $b_j$, respectively.
The `Discard Large' boxes mostly with trees and buildings are directly removed. The `Discard Small' boxes contain both potential foreground objects and background objects. 
We then apply class-agnostic tracking to associate the small background objects with foreground trajectories, and enhance the consistency of objects' sizes by using temporal coherency.

\begin{figure}[t]
  \centering
   \includegraphics[width=1\linewidth]{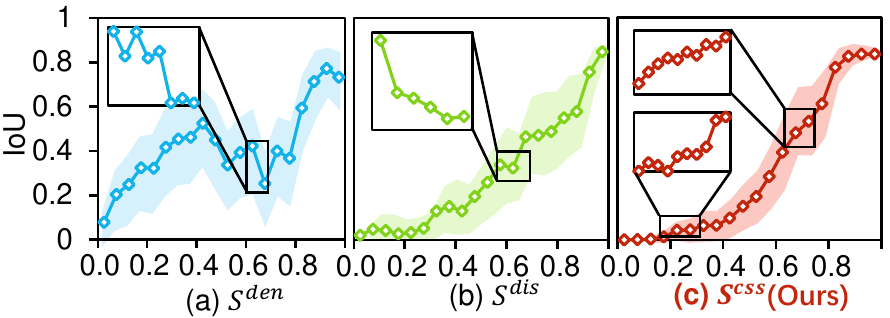}
   \caption{The comparison of different scoring methods.}
   \label{fig:sup_css_iou}
\end{figure}

\textbf{More details of CSS scoring.}
In our main paper section 3.2, we presented the CSS scoring. To better understand how the CSS scoring approximates the IoU score, we present the IoU-score carve in Fig.~\ref{fig:sup_css_iou}, where we show three methods: density scoring ($s^{den}$), distance scoring ($s^{dis}$) and our CSS scoring ($s^{css}$). Intuitively, good scoring should keep consistent with IoU scoring. In other words, with the increase of score, the selected pseudo labels should have larger IoUs with the ground truth. We found that our CSS scoring keeps the most consistent increase along with the IoU increase. Here we also provide the length, width and height of the template box for calculating the Size Similarity in the main paper Eq. 3: 
\begin{python}
{
'Vehicle': [5.06, 1.86, 1.49],
'Pedestrian': [1.0, 1.0, 2.0], 
'Cyclist': [1.9, 0.85, 1.8]
}
\end{python}

\begin{figure*}[t]
  \centering
   \includegraphics[width=1\linewidth]{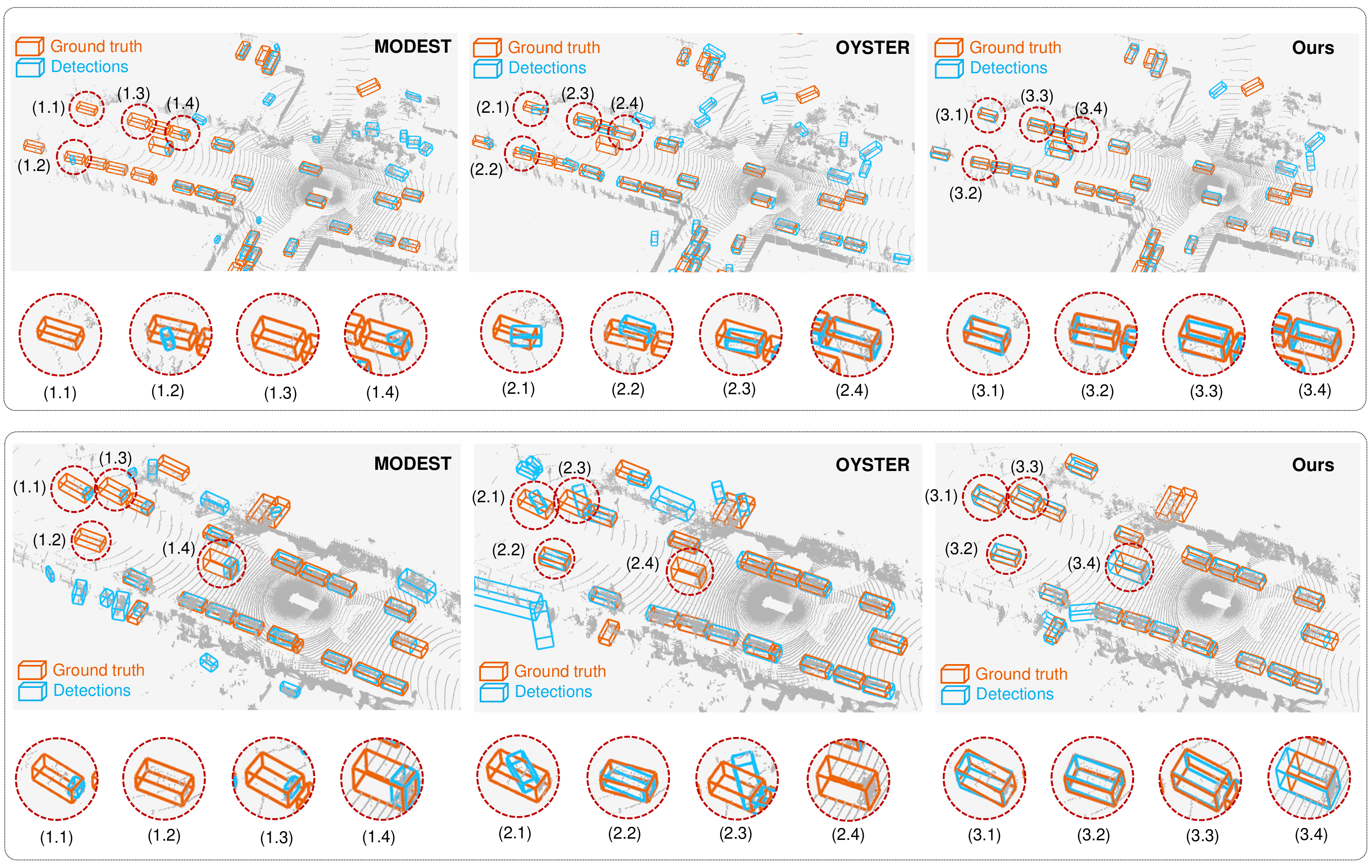}
   \caption{The visualization results predicted by different unsupervised detectors. }
   \label{fig:sup-vi}
\end{figure*}

\begin{table*}[t]
  \centering
      \resizebox{\textwidth}{!}{
    \begin{tabular}{l | c c  c c|c c c c|c c  c c}
    \hline
    \multirow{3}*{Method} & \multicolumn{4}{c|}{Vehicle} & \multicolumn{4}{c|}{Pedestrian} & \multicolumn{4}{c}{Cyclist} \\
    
                            & \multicolumn{2}{c}{3D AP L2} & \multicolumn{2}{c|}{BEV AP L2} & \multicolumn{2}{c}{3D AP L2} & \multicolumn{2}{c|}{BEV AP L2} & \multicolumn{2}{c}{3D AP L2} & \multicolumn{2}{c}{BEV AP L2} \\
    
                            & $IoU_{0.5}$ & $IoU_{0.7}$ & $IoU_{0.5}$ & $IoU_{0.7}$& $IoU_{0.3}$ & $IoU_{0.5}$& $IoU_{0.3}$ & $IoU_{0.5}$& $IoU_{0.3}$ & $IoU_{0.5}$& $IoU_{0.3}$ & $IoU_{0.5}$\\
    
    \hline
    DBSCAN & 1.94  & 0.25  & 3.97  & 1.44  & 0.19  & 0     & 2.07  & 0     & 0.2   & 0     & 0.25  & 0.06 \\

    DBSCAN+init-train & 14.87 & 2.29  & 20.6  & 11.95 & 1.35  & 0     & 6.49  & 0.1   & 0.43  & 0.2   & 0.73  & 0.24 \\

    MODEST & 15.83 & 5.48  & 19.63 & 13.31 & 8.96  & 0.1   & 14.06 & 0.13  & 1.17  & 1.01  & 2.38  & 1.07 \\

    OYSTER & 26.21 & 14.6  & 32.31 & 25.04 & 3.52  & 0.14  & 11.76 & 0.3   & 1.24  & 0.32  & 1.65  & 0.33 \\

    Proto-vanilla & 31.58 & 18.36 & 34.91 & 28.88 & 14.62 & 8.59  & 17.94 & 15.9  & 3.8   & 3.31  & 4.05  & 3.48 \\
    \hline
    \textbf{CPD(Ours)} & \textbf{50.67} & \textbf{32.13} & \textbf{52.66} & \textbf{47.48} & \textbf{20.01} & \textbf{15.22} & \textbf{20.21} & \textbf{17.26} & \textbf{5.61} & \textbf{4.87} & \textbf{5.68} & \textbf{5.22} \\
    \hline
    \end{tabular}%
    }
        \caption{3D AP L2 and BEV AP L2 results on WOD validation set.}
  \label{tab:sup-bev}%
\end{table*}%

\begin{table*}[t]
  \centering
      \resizebox{\textwidth}{!}{
    \begin{tabular}{l | c c  c c|c c c c|c c  c c}
    \hline
    \multirow{3}*{Method} & \multicolumn{4}{c|}{Vehicle 3D APH} & \multicolumn{4}{c|}{Pedestrian 3D APH} & \multicolumn{4}{c}{Cyclist 3D APH} \\
    
                            & \multicolumn{2}{c}{L1} & \multicolumn{2}{c|}{L2} & \multicolumn{2}{c}{L1} & \multicolumn{2}{c|}{L2} & \multicolumn{2}{c}{L1} & \multicolumn{2}{c}{L2} \\
    
                            & $IoU_{0.5}$ & $IoU_{0.7}$ & $IoU_{0.5}$ & $IoU_{0.7}$& $IoU_{0.3}$ & $IoU_{0.5}$& $IoU_{0.3}$ & $IoU_{0.5}$& $IoU_{0.3}$ & $IoU_{0.5}$& $IoU_{0.3}$ & $IoU_{0.5}$\\
    \hline
    MODEST & 16.43 & 4.25  & 14.04 & 3.63  & 5.59  & 0.11  & 4.18  & 0.05   & 1.07  & 0.82  & 0.45  & 0.07 \\

    OYSTER & 28.56 & 12.87 & 25.01 & 12.54 & 3.12  & 0.12  & 2.03  & 0.06  & 0.87  & 0.24  & 0.82  & 0.21 \\

    Proto-vanilla & 32.34 & 19.2  & 29.71 & 16.23 & 9.12  & 6.3   & 8.12  & 5.26  & 2.84  & 2.51  & 2.73  & 2.42 \\
    \hline
    \textbf{CPD(Ours)} & \textbf{54.19} & \textbf{34.97} & \textbf{46.99} & \textbf{30.09} & \textbf{12.01} & \textbf{9.24} & \textbf{10.06} & \textbf{7.68} & \textbf{3.68} & \textbf{3.26} & \textbf{3.55} & \textbf{3.14} \\
    \hline
    \end{tabular}%
    }
        \caption{3D APH results on WOD validation set.}
  \label{tab:sup-aph}%
\end{table*}%

\section{More Experimental Results}
\textbf{More visualization results.}
To better understand how our method improves detection results, here we present more visualization results. From Fig.~\ref{fig:sup-vi}, we observe that both the recognition and localization performance of our method (3.1-3.4) are much better than previous methods(1.1-1.4, 2.1-2.4), thanks to our CProto-based design.

\textbf{BEV AP and 3D APH results on WOD validation set.}
Some fully supervised methods also reported the BEV AP L2 and 3D APH performance. Here we presented the results in Table~\ref{tab:sup-bev} and Table~\ref{tab:sup-aph}, respectively. 
Our CPD outperforms the previous MODEST and OYSTER in both BEV AP L2 and APH L2 by a large margin, further demonstrating the effectiveness of our method.

\end{document}